\documentclass{article}
\usepackage{spconf,amsmath,graphicx}

\usepackage{float}
\usepackage{amsmath,graphicx,amsfonts,amssymb,url}
\usepackage{setspace}
\usepackage[caption=false]{subfig}
\usepackage{blindtext}
\usepackage[inline]{enumitem}
\usepackage{color}
 \usepackage{subfig}
\usepackage{multirow}
\newsavebox{\measurebox}
\usepackage{minibox}
\usepackage{array,multirow,booktabs}
\usepackage{graphicx}
\usepackage{siunitx}
\usepackage{amsfonts,amssymb}
\usepackage{balance}
\usepackage{pifont} 
\newcommand{\squeezeup}{\vspace{-2.5mm}}
\newcommand{\squeezeupf}{\vspace{-1.5mm}}
\usepackage{algorithm}
\usepackage{algorithmic}
\usepackage{url}

\newcommand*{\affaddr}[1]{#1} 
\newcommand*{\affmark}[1][*]{\textsuperscript{#1}}

\usepackage{color}

\title{Considering user agreement in learning to predict the aesthetic quality}
\name{Suiyi Ling\affmark[1], Andr\'eas Pastor\affmark[1], Junle Wang\affmark[2], Patrick Le Callet\affmark[1]  }
\address{\affaddr{\affmark[1] LS2N, \ University of Nantes} \ \  \affaddr{\affmark[2]Turing Lab, \  Tencent }}

\begin{document}
 
\maketitle
 
\begin{abstract}
How to robustly rank the aesthetic quality of given images has been a long-standing ill-posed topic. Such challenge stems mainly from the diverse subjective opinions of different observers about the varied types of content. There is a growing interest in estimating the user agreement by considering the standard deviation ($\sigma$) of the scores, instead of only predicting the mean aesthetic opinion score ($\mu$). Nevertheless, when comparing a pair of contents, few studies consider how confident are we regarding the difference in the aesthetic scores. In this paper, we thus propose (1) a re-adapted multi-task attention network to predict both the mean opinion score and the standard deviation in an end-to-end manner; (2) a brand-new confidence interval ranking loss that encourages the model to focus on image-pairs that are less certain about the difference of their aesthetic scores. With such loss, the model is encouraged to learn the uncertainty of the content that is relevant to the diversity of observers’ opinions, \textsl{i.e.}, user disagreement. Extensive experiments have demonstrated that the proposed multi-task aesthetic model achieves state-of-the-art performance on two different types of aesthetic datasets, \textsl{i.e.}, AVA and TMGA. 
\end{abstract}

\squeezeup \squeezeupf 
\section{Introduction}
\label{sec:intro} \squeezeup 
Visual aesthetic assessment is important for many real-world use cases, including automatic image composition~\cite{su2021camera}, image creation, editing~\cite{xu2020spatial}, graphical design \textsl{etc.} and is intriguing more other piratical research topics. As people differ in how they respond to artworks~\cite{schlotz2020aesthetic}, unlike general quality assessment, aesthetic assessment is associated more with high-level components of the contents in terms of emotions, composition, and beauty. Thus, it is more subjective compared to quality assessment of compressed/distorted contents~\cite{talebi2018nima}. State-of-The-Art (SoTA) approaches~\cite{pfister2021self,xu2020spatial,hosu2019effective} concentrate on leveraging different deep neural networks with large-scale annotated aesthetic data, and achieve decent performance on benchmark datasets in recent yet. Other than predicting the atheistic quality score in terms of mean opinion scores ($\mu$) alone, more and more studies, \textsl{e.g.} NIMA~\cite{talebi2018nima}, developed their models by taking the standard deviation ($\sigma$) of the scores, \textsl{i.e.}, how observer agree with each other, into account.

\begin{figure}[t]
    \centering
    \includegraphics[width=\columnwidth]{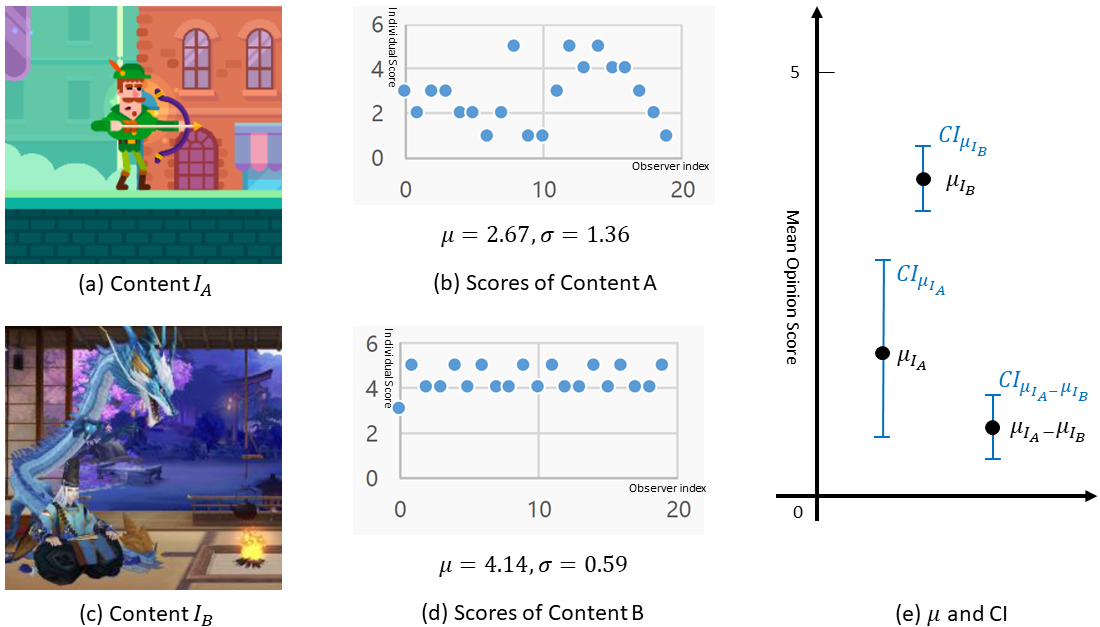}
    \caption{Example of contents that have significantly different standard deviation $\sigma$, and demonstration of how Confidence Intervals (CI) could help to rank the aesthetic scores. \squeezeupf }  
    \label{fig:teaser} \squeezeup 
\end{figure}

When ranking the aesthetic qualities of a pair of contents, it is important to not only comparing their mean opinions scores, but also the corresponding confidence intervals. For instance, as shown in Fig~\ref{fig:teaser}, as the art-style of image $I_A$ is simpler than $I_B$, its aesthetic votes are polarized. As depicted in Fig~\ref{fig:teaser} (b), some of the observers voted very high score for $I_A$ as they preferred the style, while most of the other observers gave very low scores. As a result, the standard deviation of its aesthetic score $\sigma_{I_A}$ is much higher than the one of $I_B$. When comparing the aesthetic quality between $I_A$ and  $I_B$, we want to make a statistically solid decision based on the opinions of the majority. In other words, as presented in Fig~\ref{fig:teaser} (e), we consider $I_B$ is significantly better than $I_A$ only if the confidence intervals of the two contents are not overlapped~\cite{tiotsop2021modeling,ling2020strategy}. In general, the Confidence Interval (CI) of the aesthetic score of a given image $I_{A}$ is given by:  \squeezeupf 
 \begin{equation}
 CI_{\mu_{I_A}} = \mu_{I_A}\pm z \frac{\sigma_{I_{A}}}{\sqrt{n_{obs}}} \squeezeupf 
 \end{equation}
 where $z =	1.96$ is the critical value \textsl{w.r.t.} $95\%$ of confidence level, and $n$ is the number of observers. However, within modern deep learning framework, it is impractical to  compute first the CIs and then compare them (\textsl{e.g.}, limited by the design of loss function). Therefore, as an alternative, we may prefer to know how confident we are regarding the difference of the aesthetic scores of a given pair. If $CI_{\mu_{I_{A}} - \mu_{I_{B}}}$ is small enough, we could then say confidently that they are significantly different, where $CI_{\mu_{I_{A}} - \mu_{I_{B}}}$ is the confidence interval of $\mu_{I_{A}} - \mu_{I_{B}}$, defined as:     \squeezeupf    
\begin{equation}
CI_{\mu_{I_{A}} - \mu_{I_{B}}}= |\mu_{I_{A}} - \mu_{I_{B}}| \pm z \cdot \sigma_{\mu_{I_{A}} - \mu_{I_{B}}}
\label{eq_CI_Dif} \squeezeupf  
\end{equation}
It is not hard to see that the standard deviations of the observers' scores $\sigma$ are in both aforementioned equations, and thus are important to be predicted along with the mean opinion score $\mu$. Based on the discussion above, the contribution of this study is two-fold: (1) we propose to predict both $\mu$ and $\sigma$ together via a novel multi-task attention network; (2) inspired by equation~\eqref{eq_CI_Dif}, we develop a new confidence interval loss so that the proposed model is capable of learning to rank the aesthetic quality regarding the confidence intervals.

\squeezeup  \squeezeupf
\section{Related Work}   \squeezeup   
In the past decade, the performances of aesthetic assessment models grow at a respectable pace. Li \textsl{et al.}~\cite{li2009aesthetic} proposed one of the early efficient aesthetic metrics based on hand-crafted features. By formulating aesthetic quality assessment as a ranking problem, ~\cite{kao2015visual}, Kao \textsl{et al.} developed a rank-based methodology for aesthetic assessment. Akin to~\cite{kao2015visual},  another ranking network was proposed in~\cite{kong2016photo} with attributes and content adaptation. To facilitate heterogeneous input, a double-column deep network architecture was presented in~\cite{lu2015rating}, which was improved subsequently in~\cite{lu2015deep} with a novel multiple columns architecture. Ma \textsl{et al.} developed a salient patch selection approach~\cite{ma2017lamp} that achieved significant improvements. Three individual convolutional neural networks (CNN) that capture different types of information were trained and integrated into one final aesthetic identifier in~\cite{kao2016hierarchical}. Global average pooled activations were utilized by Hii~\textsl{et al.} in~\cite{hii2017multigap} to take the image distortions into account. Later, triplet loss was employed in a deep framework in~\cite{schwarz2018will} to further push the performances to the limits of most modern methods available at the time. The Neural IMage Assessment (NIMA)~\cite{talebi2018nima}, developed by Talebi~\textsl{et al.}, is commonly considered as the baseline model. It was the very first metric that evaluates the aesthetic score via predicting the distribution of the ground truth data. To assess UAV video aesthetically, a deep multi-modality model was proposed~\cite{kuang2019deep}. As global pooling is conducive to arbitrary high-resolution input, MLSP~\cite{hosu2019effective} was proposed, based on Multi-Level Spatially Pooled features (MLSP). Recently, an Adaptive Fractional Dilated Convolution (AFDC)~\cite{chen2020adaptive} was proposed to incorporate the information of image aspect ratios.

\squeezeup \squeezeupf   
\section{The proposed Model} \squeezeup  
\subsection{Deep Image Features for Arbitrary Input}
\begin{figure}[!htbp]
    \centering
    \includegraphics[width=\columnwidth]{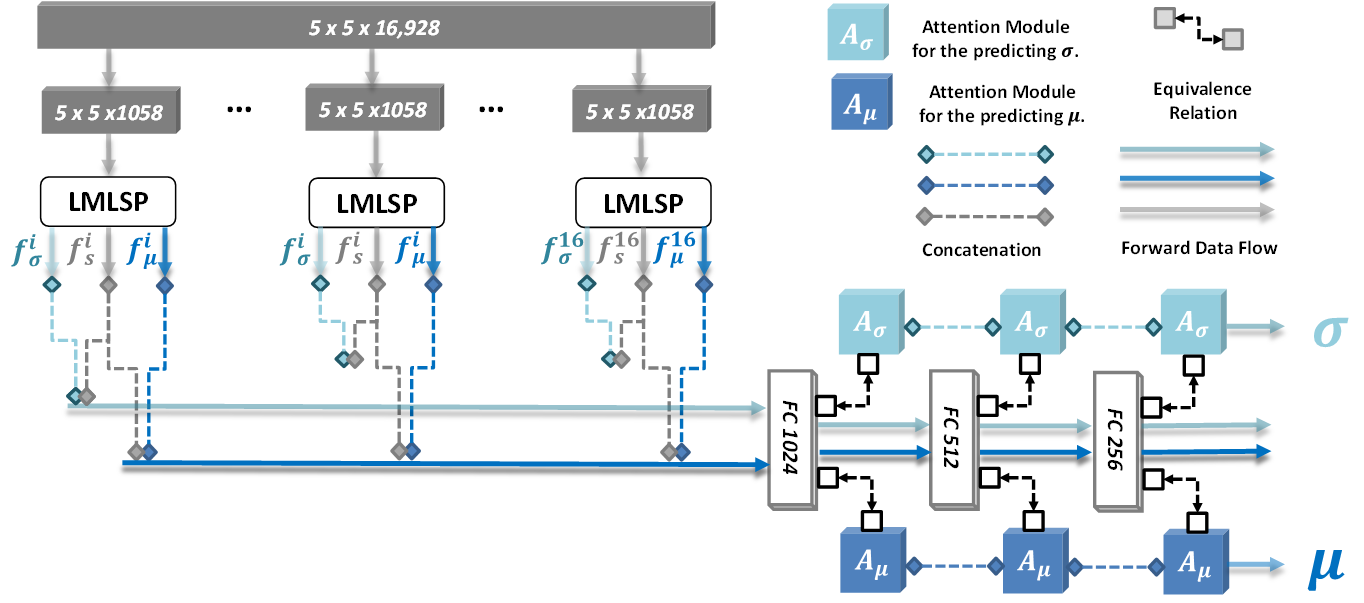} \squeezeupf 
    \caption{Overall network architecture of the proposed model. \squeezeup}  
    \label{fig:overall}
\end{figure}

\begin{figure}[!htbp]
    \centering
    \includegraphics[width=0.7\columnwidth]{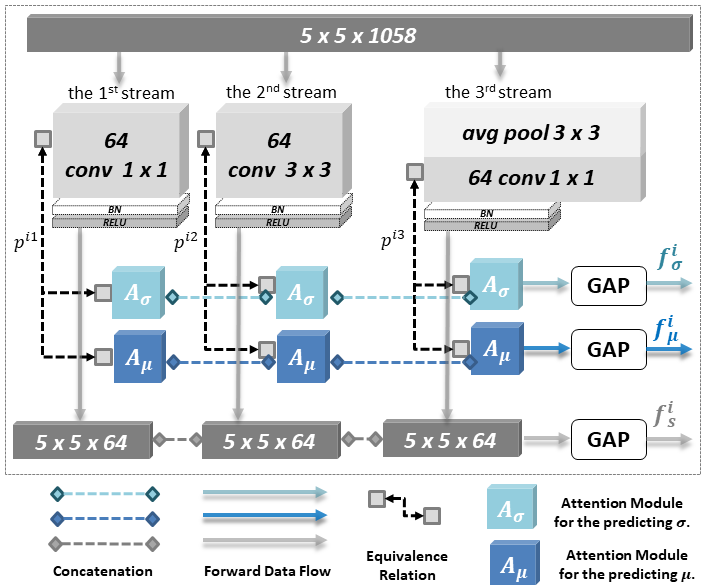}
    \caption{Detailed visualization of each LMLSP block. \squeezeup}
    \label{fig:LMLSP}
\end{figure}

\squeezeupf   
It is verified in~\cite{hosu2019effective} that the wide MLSP feature architecture~\textsl{Pool-3FC} achieves the SoTA performance. Thus, it is adopted in this study as the baseline model. Similar to MLSP, we extracted features from the Inception ResNet-v2 architecture. As the dimension of the features, \textsl{i.e.}, $5 \times 5 \times 16928$ is too high, the features are first divided into 16 sub-features with dimension of $5 \times 5 \times 1058$ (\textsl{i.e.}, the dark-grey blocks at left-upper corner of Fig.~\ref{fig:overall}) to ease the computation of attention feature and reduce the number of parameters used across the architecture. After the division, each sub-feature is fed into a Lighter-weight Multi-Level Spatially Pooled feature block (LMLSP) as depicted in Fig.~\ref{fig:overall}. The module of LMLSP is shown in Fig.~\ref{fig:LMLSP}. It is composed of three streams, including the $1\times 1$, $3\times 3$ convolutions streams, and the stream of average pooling followed by a $1 \times 1$ convolution. Through this design, the dimension of the multi-level spatial pooled feature was reduced to facilitate the latter fully connected layers.

 \squeezeupf 
 \squeezeup 
 \squeezeup   \squeezeup   \squeezeupf
\subsection{Predicting  $\mu$ and $\sigma$ at the Same Time with Multi-Task Attention Network}

The goal of using multi-task learning is two-fold, \textsl{i.e.}, (1) predict both the mean opinion scores and the standard deviation at the same time; (2) via joint loss functions, the network learns to consider observers' diverse opinions ($\sigma$), the content uncertainty, \textsl{etc.}, when predicting the final aesthetic score. 

In this study, the multi-task attention network proposed in~\cite{liu2019end} is adapted to predict both the mean opinion score $\mu$, and the standard deviation of the subjects' scores $\sigma$. The proposed LMLSP blocks are utilized as task-shared features blocks. And the separate attention modules, which are linked to the shared model, are designed to learn task-specific features for $\mu$ and $\sigma$ correspondingly. The attention modules of $\mu$ (in cyan color) and $\sigma$ (in dark blue color) are shown in Fig.~\ref{fig:overall} and~\ref{fig:LMLSP}. As presented in Fig.~\ref{fig:LMLSP}, within each individual LMLSP block $i$, two attention masks are attached to each convolution stream, \textsl{i.e.}, one for task $\mu$ and one for task $\sigma$ at each conv $1\times 1$, conv $3\times 3$, and the average pooling stream. The task specific features $ \hat{a}_{\sigma}^{ij}$ and $ \hat{a}_{\mu}^{ij}$ in the $i_{th}$ LMLSP block, at the $j_{th}$ stream are calculated via element-wise multiplication of the corresponding attention modules $A_{\sigma}/A_{\mu}$  with the shared stream features $p^{ij}$:  \squeezeup  
\begin{equation}
 \hat{a}_{\sigma}^{ij} = A_{\sigma} \odot p^{ij},  i \in[1,16], j \in [1,3] \squeezeupf
\label{eq:a_sigma}
\end{equation} 
\begin{equation}
 \hat{a}_{\mu}^{ij} = A_{\mu} \odot p^{ij}, i \in[1,16], j \in [1,3]
\label{eq:a_mu}
\end{equation}
 where $\odot$ indicates the element-wise multiplication operation. 

Then, for each task $\mu$/$\sigma$, the task-specific features obtained from each stream are concatenated into one dedicated feature, followed by a Global Average Pooling
(GAP). Same for the shared features from the three streams. As such, for the $i_{th}$ LMLSP block, three features are output, including $f_{\mu}^i$, $f_{\sigma}^i$ and $f_{s}^i$, which are the feature of task $\mu$, the feature of task $\sigma$ and the feature of the shared network respectively:   \squeezeup   
\begin{equation}
f_{\mu}^i = GAP( [\hat{a}_{\mu}^{i1}, \hat{a}_{\mu}^{i2}, \hat{a}_{\mu}^{i3} ] ), i \in[1,16] \squeezeupf     
\end{equation} 
\begin{equation}
f_{\sigma}^i = GAP( [\hat{a}_{\sigma}^{i1}, \hat{a}_{\sigma}^{i2}, \hat{a}_{\sigma}^{i3} ] ),  i \in[1,16]       
\end{equation}
\begin{equation}
f_{s}^i = GAP( [f_{\sigma}^{i1}, f_{\sigma}^{i2}, f_{\sigma}^{i3} ] ),  i \in[1,16],  
\end{equation}
where $GAP(\cdot)$ is the Global Average Pooling and $[ X, Y]$ denotes the concatenation of tensor $X$ and $Y$.

Afterward, as depicted in the lower part of Fig.~\ref{fig:overall}, for each task, \textsl{i.e.}, $\mu$, or $\sigma$, at each LMLSP block, the task-specific feature $f_{\mu/\sigma}^i$ is first concatenated with the shared feature $f_{s}^i$, and all the obtained features are further concatenated across all the 16 LMLSP blocks to generate the final feature for each task. The feature flow is highlighted in cyan color for task $\sigma$ and dark blue color for task $\mu$. Lastly, the features are forwarded to three continuous Fully-Connected (FC) layers with the same attention modules for each task after each FC layer to predict the final $\mu$ and $\sigma$. It is worth mentioning that the entire network is trained in an end-to-end manner. As such, the attention modules for $\mu$, and $\sigma$ serves as feature selectors that pick up the relevant dimensions in predicting $\mu$ or $\sigma$ respectively, while the shared LMLSP blocks and the shared FC layers learn the general features across the two tasks.  

\squeezeup \squeezeupf  
\subsection{Loss function} \squeezeup 
Under the multi-task learning setting, the joint loss function $L_{MTL}$ of the tasks of predicting the mean $\mu$ and the standard deviation of subject' scores is defined as: \squeezeup
\begin{equation}
\mathcal{L_{MTL}} = \alpha_{\mu} \mathcal{L_{\mu}} +  \alpha_{\sigma} \mathcal{L_{\sigma}},
\label{eq:mtl}
\end{equation}
where $\alpha_{\mu}$ and $\alpha_{\sigma}$ are the parameters that balances the losses of predicting $\mu$, \textsl{i.e.}, $\mathcal{L_{\mu}}$ and the one of predicting $\sigma$ \textsl{i.e.}, $\mathcal{L_{\sigma}}$. The $\mathcal{L_{\sigma}}$ is simply defined as the Mean Absolute Error (MAE) between the ground truth $\sigma_{I}$ of an image $I$, and the corresponding predicted   $\hat{\sigma}_{I}$. N equals to the number of images:
\begin{equation}
L_{\sigma} = \frac{\sum^N_{I=1} | \sigma_{I} - \hat{\sigma}_{I} |}{N} 
\end{equation}

As emphasized in Section~\ref{sec:intro}, when comparing the aesthetic scores of two images $I_{A}$ and $I_{B}$, we want to know not only whether image $I_{A}$ is significantly better than $I_{B}$ in terms of their aesthetic quality, but also how confident we are regarding the difference. Specifically, we want to consider this `certainty' when predicting $\mu$. Thus, the ${L_{\mu}}$ is defined with a novel $\mathcal{L}_{CI}$, namely, the~\textsl{Confidence Interval Loss}:
\begin{equation}
\mathcal{L_{\mu}} = (1-\lambda) \frac{\sum^N_{I=1} | \mu_{I} - \hat{\mu}_{I} |}{N}  + \lambda \cdot \mathcal{L}_{CI}, 
\label{eq:overall}
\end{equation}
where $\lambda$ is a parameter that balances the two losses. Inspired by the confidence interval of the differences between two aesthetic scores $CI_{\mu_{I_{A}} - \mu_{I_{B}}}$, which is defined as  $|\mu_{I_{A}} - \mu_{I_{B}} | \pm z \cdot \sigma_{\mu_{I_{A}} - \mu_{I_{B}}}$ (shown in Section~\ref{sec:intro}), we further define $\mathcal{L}_{CI}$ as:  \squeezeup
 \begin{equation}
\begin{split}
\mathcal{L}_{CI} = & \sum_{ (I_{A}, I_{B})  \subset  S_{pair} }  max\big(0, \\ 
& \ l_{CI}(\mu_{I_{A}},\mu_{I_{B}}) \cdot [ \lvert \   |\mu_{I_{A}}-\mu_{I_{B}}| - |\hat{\mu_{I_{A}}} -  \hat{\mu_{I_{B}}}| \ \rvert]  \big), \\
\end{split}
\label{eq_CI}
\end{equation}
where $\mu_{I}$ and $\hat{\mu_{I}}$ are the ground truth mean and the predicted mean respectively, $ S_{pair}$ is the set of all possible pairs, and $l_{CI}$ is defined as below: \squeezeup   
\begin{equation}
l_{CI}= \left\{ \begin{array}{rcl}
 1, \  z \cdot \sigma_{\mu_i - \mu_j}  > \tau \\  
 0,    \ \ \ \ otherwise \ \ \ 
\end{array}\right.
\label{eq:lci_1}
\end{equation}
where $\tau$ is the margin. Without loss of generality, $\sigma_{y_i - y_j} \approx \sqrt{ \frac{\sigma_i^2}{n_{obs}}  + \frac{\sigma_j^2}{n_{obs}} }$~\cite{rumsey2009statistics}. $n_{obs}$ indicates the number of observers. $l_{CI}$ serves as a gating function. In equation \eqref{eq_CI}, the prediction error of a given image pair is accumulated only if $l_{CI}$ equals to 1, under the condition of $z \cdot \sigma_{\mu_i - \mu_j}  > \tau$. It can be noticed that, $z \cdot \sigma_{\mu_i - \mu_j}$ is from the \eqref{eq_CI_Dif}, \textsl{i.e.}, the definition of $CI_{\mu_{I_{A}} - \mu_{I_{B}}}$. The proposed~\textsl{Confidence Interval Loss} punishes only the pair that have $CI_{\mu_{I_{A}} - \mu_{I_{B}}}$ value that higher than the margin $\tau$. In other words, it focuses on pairs that have higher uncertainty (larger confidence interval of the aesthetic score difference). Furthermore, the loss for pair that triggers the gate function is $[|\mu_{I_{A}}-\mu_{I_{B}}| - |\hat{\mu_{I_{A}}} -  \hat{\mu_{I_{B}}}| ]$, which encourages the prediction of the difference between the predicted scores to be as close as the ground truth one. 

The intuition behind this selection of pairs is to concentrate on pairs that have larger uncertainties, which further enables the model to learn uncertainty from the ambiguous contents and improve the performance of the model on such pairs.

\squeezeup \squeezeupf
\section{Experiment}
\squeezeup
\subsection{Experimental setup} \squeezeup 
The performance of the proposed model was assessed on two different types of aesthetic quality datasets that were developed for different use cases/scenarios:

(1) The Tencent Mobile Gaming Aesthetic (TMGA) dataset~\cite{ling2020subjective}, which is developed for improving gaming aesthetics. In this dataset, there are in total 1091 images collected from 100 mobile games, where each image was labeled with four different dimensions including the `Fineness', the `Colorfulness', the `Color harmony', and the `Overall aesthetic quality'. The entire dataset is divided into 80\%, 10\%, and 10\%, for training, validation, and testing correspondingly.

(2) The Aesthetic Visual Analysis (AVA) dataset~\cite{murray2012ava}, which was developed for general aesthetic purposes. It contains about 250 thousand images that were collected from a photography community. Each individual image was rated by an average of 210 observers. The same random split strategy was adapted as done
in~\cite{chen2020adaptive,hosu2019effective} to obtain 235,528 images for training/validation (95\%/5\%) and 20,000 for test. 


Since most of the metrics were evaluated on the AVA dataset regarding the binary classification performance, ACCuracy (ACC) was computed based on classifying images into two classes with a cut-off score equals to 5~\cite{talebi2018nima,hosu2019effective} when reporting the performances on the AVA dataset. However, as emphasized in~\cite{hosu2019effective},   this predominant two-class accuracy suffers from several pitfalls. For instance, due to the unbalanced distribution of images in training, testing set (different numbers of images of different aesthetic quality score values), using `accuracy' does not fully reveal/stress out the performances of under-test metrics regarding its capability in ranking the aesthetic score of the image. Thus, as done in ~\cite{hosu2019effective,ling2020subjective}, we computed the Pearson correlation coefficient (PCC), and the Spearman’s rank order correlation coefficient (SCC) between the ground truth and the predicted $\mu$ to evaluate the performance of different aesthetic quality assessment approaches. Similar to~\cite{talebi2018nima}, as the range of $\sigma$ differs across the dataset, the performances of models in terms of predicting $\sigma$ are reported in PCC, and SCC. The proposed model is mainly compared to SoTA aesthetic models AFDC~\cite{chen2020adaptive}, MLSP~\cite{hosu2019effective} and one of the most popular baseline models NIMA~\cite{talebi2018nima}.

During training, we applied ADAM optimizer with an initial learning rate of $10^{-4}$, which was further divided by 10 every 20 epochs. As a preliminary study to learn both $\sigma$ and $\mu$ with multi-task attention network by taking confidence interval into account, we still want to focus on predicting $\mu$. Thus, the $\lambda_{\mu}$ and $\lambda_{\sigma}$ in~\eqref{eq:mtl} were set as 0.6 and 0.4 accordingly, with slightly higher weight for task $\mu$. $\lambda$ in \eqref{eq:overall} was set as 0.5.

\squeezeup \squeezeupf
\subsection{Experimental Results} \squeezeup
\textbf{Results on AVA dataset:} The PCC, SCC, and ACC values of the considered models in predicting $\mu$ and $\sigma$ on the AVA dataset are shown in Table 1. Regarding task $\mu$, the proposed model accomplishes state-of-the-art performance by obtaining the best PCC, SCC values, and second best ACC value. Regarding task $\sigma$~\footnote{As the code of `AFDC' is not publicly available, its performance in predicting $\sigma$ on AVA, and its performances on TMGA are not reported.}, the proposed model achieves the best performance. It can be noticed that the performances of objective models in predicting $\sigma$ are lower than the ones of $\mu$. One of the potential reasons is that the task of predicting $\sigma$ is more challenging than $\mu$, as it is not only relevant to the contents but also the subjects who participated in the test.

\textbf{Results on TMGA dataset:} 
For a fair comparison,  NIMA and MLSP were first finetuned on the training set of TMGA dataset with the optimized hyper-parameters. As the performance of MLSP is higher than NIMA by a large margin, we only report the results of MLSP. The results are presented in Table~\ref{tab:res_TMGA}. Similarly, the performances of our model are superior to the ones of the compared models.   \squeezeup    \squeezeup   
\begin{table}[!htbp]
\begin{center}
\centering
\footnotesize
\caption{Performances of relevant models~\textbf{AVA} dataset. \squeezeup \squeezeupf} \squeezeup   
{\renewcommand{\baselinestretch}{1}
\begin{tabular}{|c|c|c|c|c|c|c|}
\hline
&\bf{PCC ($\mu$)} &\bf{SCC ($\mu$)} &   \bf{ACC}  & \bf{PCC ($\sigma$)} & \bf{SCC($\sigma$)}    \\ \hline
 
NIMA~\cite{talebi2018nima} & 0.636 & 0.612 &  81.51\% & 0.233 & 0.218  \\ 
MLSP~\cite{hosu2019effective} & 0.756  & 0.757 &    81.72\% & 0.568  &  0.549    \\
AFDC~\cite{chen2020adaptive} & 0.671   & 0.648   &   \textbf{83.24\%}   & - & -    \\ 
\textbf{Proposed}  &  \textbf{0.785}  &   \textbf{0.779} &  82.97\% & 0.624 & 0.613  \\ \hline
\end{tabular}}
\end{center}
\squeezeup    
\label{tab:ava}
\end{table}

\squeezeup   \squeezeup    \squeezeup   \squeezeup   
\begin{table}[!htbp]
\footnotesize
\centering
\caption{ Performances of relevant models on~\textbf{TMGA} dataset.}
\begin{tabular}{|c|cccc|}  
\hline
 & \textbf{Fineness} & \textbf{Colorful} & \textbf{Harmony} & \textbf{Overall} \\
\hline
\multicolumn{5}{|c|}{\textbf{ PCC of models in predicting $\mu$ }} \\ 
\hline
MLSP~\cite{hosu2019effective} & 0.904 & 0.900 & 0.888  & 0.872 \\ 
\textbf{Proposed} & \textbf{0.923} & \textbf{0.936} & \textbf{0.918} & \textbf{0.910} \\
\hline   
\multicolumn{5}{|c|}{\textbf{SCC of models in predicting $\mu$}} \\ \hline
MLSP~\cite{hosu2019effective} & 0.904 & 0.904 & 0.826 & 0.8652 \\
\textbf{Proposed} & \textbf{0.917} & \textbf{0.923} & \textbf{0.889} & \textbf{0.897} \\    \hline
\multicolumn{5}{|c|}{\textbf{ PCC of models in predicting $\sigma$ }} \\ \hline
MLSP~\cite{hosu2019effective} & 0.607 & 0.413  & 0.289  &  0.5151  \\ 
Proposed & \textbf{0.686} & \textbf{0.557} & \textbf{0.372} & \textbf{0.559} \\   \hline   
\multicolumn{5}{|c|}{\textbf{SCC of models in predicting $\sigma$ }} \\ \hline
MLSP~\cite{hosu2019effective}  &  0.555  & 0.364 & 0.301  & 0.475 \\
Proposed & \textbf{0.679} & \textbf{0.486} & \textbf{0.370} & \textbf{0.478}  \\    \hline
\end{tabular}
\label{tab:res_TMGA} \squeezeup    
\end{table} 

\squeezeup    \squeezeupf    
\subsection{Ablation Study}  \squeezeup  
To validate the effeteness of each component within the proposed model, ablation studies (with 2 ablative models) were conducted. The results are presented in Table 3: (1) when comparing rows 1 and 2, it is demonstrated that by employing the proposed $\mathcal{L}_{CI}$ the performance of the baseline single-task MLSP model in predicting $\mu$ is improved; (2) By comparing row 1 and 3, it is showcased that the performance of the aesthetic model is boosted by learning both $\sigma$ and $\mu$ together via multi-task attention network.

\squeezeup  \squeezeup  
\begin{table}[!htbp]
\begin{center}
\caption{
Ablation studies on the \textbf{AVA} dataset. }
{
\renewcommand{\baselinestretch}{1}\footnotesize
\begin{tabular}{|c|c|c|c|c|}
\hline
&\bf{PCC} &\bf{SCC} &  \bf{ACC}     \\ \hline
Single-Task $\mu$ (MLSP) ~\cite{hosu2019effective} & 0.756 & 0.757 & 81.72\%   \\
Single-Task $\mu$ with $\mathcal{L}_{CI}$ & 0.762 & 0.765  &   81.89\% \\
Multi-Task $\mu$ \& $\sigma$ without  $\mathcal{L}_{CI}$ & 0.775  & 0.772   &  82.24\%\\
Multi-Task $\mu$ \& $\sigma$ with $\mathcal{L}_{CI}$ \textbf{(Proposed)}  &  \textbf{0.785}  &   \textbf{0.779} &  \textbf{82.97\%}  \\ \hline
\end{tabular}}
\end{center}
\label{tab:ablation} \squeezeup    \squeezeup   
\end{table}
 \squeezeup  \squeezeup  \squeezeupf
\subsection{Conclusion}
 In this study, a novel aesthetic model is proposed to learn the aesthetic ranking by exploiting the confidence intervals based on (1) learning both the mean and the standard deviation of aesthetic scores through multi-task attention network; (2) a new confidence interval loss that facilitates the model to concentrate on the less confident pairs and learn the high-level ambiguous characteristics of the contents. Through experiments, it is showcased that our model achieves superior performances compared to the state-of-the-art models in predicting both the mean and the standard deviation of aesthetic scores.  
\newpage 
\balance 
\small{
\bibliographystyle{IEEE}
\bibliography{ref}
}
\end{document}